\pdfoutput=1

\documentclass[11pt]{article}

\usepackage[final]{acl}

\usepackage{times}
\usepackage{latexsym}

\usepackage[T1]{fontenc}

\usepackage[utf8]{inputenc}

\usepackage{microtype}

\usepackage{inconsolata}
\usepackage{url}
\usepackage{graphicx}
\usepackage{multirow}
\usepackage{amssymb}
\usepackage{amsmath}
\usepackage{booktabs}
\usepackage{pifont}%
\usepackage{subcaption}
\newcommand{\cmark}{\ding{51}}%
\newcommand{\xmark}{\ding{55}}%

\title{T2S-GPT: Dynamic Vector Quantization for Autoregressive Sign Language Production from Text}

\author{
Aoxiong Yin,
Haoyuan Li,
Kai Shen,
Siliang Tang\thanks{Corresponding author.},
Yueting Zhuang \\
\small Zhejiang University\\
\tt\small \{yinaoxiong,lihaoyuan,shenkai,siliang,yzhuang\}@zju.edu.cn
}

\begin{document}
\maketitle
\begin{abstract}
In this work, we propose a two-stage sign language production (SLP) paradigm that first encodes sign language sequences into discrete codes and then autoregressively generates sign language from text based on the learned codebook.
However, existing vector quantization (VQ) methods are fixed-length encodings, overlooking the uneven information density in sign language, which leads to under-encoding of important regions and over-encoding of unimportant regions.
To address this issue, we propose a novel dynamic vector quantization (DVA-VAE) model that can dynamically adjust the encoding length based on the information density in sign language to achieve accurate and compact encoding.
Then, a GPT-like model learns to generate code sequences and their corresponding durations from spoken language text.
Extensive experiments conducted on the PHOENIX14T dataset demonstrate the effectiveness of our proposed method.
To promote sign language research, we propose a new large German sign language dataset, PHOENIX-News, which contains 486 hours of sign language videos, audio, and transcription texts.
Experimental analysis on PHOENIX-News shows that the performance of our model can be further improved by increasing the size of the training data.
Our project homepage is \url{https://t2sgpt-demo.yinaoxiong.cn}.

\end{abstract}

\section{Introduction}
Sign language is a visual language with complex grammatical structures and is the primary means of communication for nearly 70 million deaf people worldwide \footnote{According to World Federation of the Deaf \url{https://wfdeaf.org/our-work/}}.
Research on sign language production \citep{baltatzis2023neural,fangSignDiffLearningDiffusion2023,huangFastHighQualitySign2021,hwangNonautoregressiveSignLanguage2021,hwangNonAutoregressiveSignLanguage2022,saundersAdversarialTrainingMultichannel2020} and sign language translation \citep{camgoz2018neural,zhang2023sltunet,zhouImprovingSignLanguage2021,yin2021simulslt,yin2023gloss} has attracted widespread attention.
Sign language production (SLP) is a challenging problem that aims to automatically translate spoken language descriptions into corresponding continuous sign sequences. 
SLP can help deaf people better access information and communicate with others, thereby facilitating their lives, which has important social significance.

\begin{figure}
    \centering
    \includegraphics[width=\linewidth]{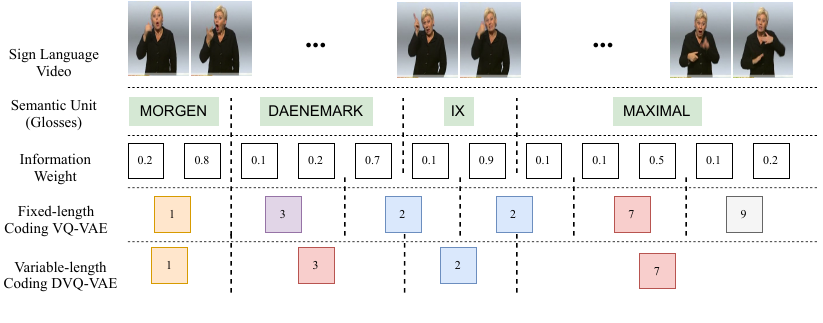}
    \caption{Comparison of fixed-length encoding and variable-length encoding.}
    \label{fig:introduction}
\end{figure}

SLP models are expected to learn precise mapping from the spoken language space to the sign language space.
Early work used 2D or 3D skeleton poses to represent sign language \citep{huangFastHighQualitySign2021, saundersMixedSIGNalsSign2021, saundersProgressiveTransformersEndtoend2020}, while recent work has suggested using 3D human models, such as SMPL-x\citep{pavlakosExpressiveBodyCapture2019}, to represent sign language, as it introduces human priors and can better animate \citep{baltatzisNeuralSignActors2023}.
To learn the mapping between these two different modal spaces, some work uses autoregressive models \citep{saundersContinuous3DMultichannel2021, saundersMixedSIGNalsSign2021, saundersProgressiveTransformersEndtoend2020}, non-autoregressive models \citep{huangFastHighQualitySign2021, hwangNonautoregressiveSignLanguage2021, hwangNonAutoregressiveSignLanguage2022}, or diffusion models \citep{baltatzisNeuralSignActors2023} to learn the direct mapping from spoken language text to sign language skeleton poses.
\citep{xieG2PDDMGeneratingSign2023} proposed to learn the discrete representation of sign language through VQ-VAE \citep{vandenoordNeuralDiscreteRepresentation2017} and then learn the mapping from text to discrete representation through a discrete diffusion model.
However, we found that existing sign language discrete representation methods are fixed-length encodings, as shown in \autoref{fig:introduction}, which overlooks the uneven information density in sign language.
In addition, many existing works rely on expert-annotated intermediate representations, i.e. glosses, which limit the scalability of the model.

\begin{table*}[ht]
  \centering
  
  \caption{ Summary statistics for different sign language datasets. }
        \label{tab:dataset}
  \resizebox{\linewidth}{!}{
  \begin{tabular}{l|c|cccc|crc|c}
  \hline
  \multirow{2}{*}{Dataset} & \multirow{2}{*}{Language} &\multicolumn{4}{c|}{Attribute}   &\multicolumn{3}{c|}{Statitics}& \multirow{2}{*}{Source} \\
         & &Transcription &Pose &Speech  &Document-level & Duration(h)   &Vocab  & Signers&            \\ 
         \hline 
  BOBSL \citep{albanie2021bbc}                           &BSL&\cmark &\cmark &\cmark &\cmark &1447  & 77k & 39 &TV   \\
  How2Sign\citep{duarteHow2SignLargescaleMultimodal2021a}&ASL&\cmark &\cmark &\cmark &\xmark & 79   &16k & 11 &Lab   \\ 
  OpenASL\citep{shiOpenDomainSignLanguage2022}          &ASL&\cmark &\xmark &\xmark &\xmark & 288   &33k & 220 &Web   \\  
  YouTube-ASL\citep{uthusYouTubeASLLargeScaleOpenDomain2023}&ASL&\cmark &\xmark &\xmark  &\cmark  &984  & 60k & >2519      &Web   \\ 
  CSL-Daily\citep{zhouImprovingSignLanguage2021} &CSL&\cmark &\xmark &\xmark  &\xmark  &23  &2k & 10       &Lab    \\ 
  SWISSTXT\citep{camgoz2021content4all} &DSGS&\cmark &\cmark &\cmark & \cmark &88   &  - &  -       &TV    \\
  VRT-RAW\citep{camgoz2021content4all} &VGT&\cmark &\cmark &\cmark & \cmark &100  &- & -         &TV    \\ 
  KETI\citep{ko2019neural}      &KVK&\cmark &\cmark&\xmark    &\xmark & 29   &419 & 14        &Lab   \\ 
  SP-10\citep{yinMLSLTMultilingualSign2022b} &various&\cmark &\cmark&\xmark  &\xmark   & 14   &17k & 79       &Web  \\ 
  AfriSign\citep{gueuwou2023afrisign} &various&\cmark &\xmark&\xmark  &\xmark   & 152   &20k & -       &Web  \\ 
  \hline  
  PHOENIX2014T\citep{camgozNeuralSignLanguage2018} &DGS&\cmark &\xmark&\xmark & \xmark   &11  &3k  & 9         &TV    \\  
  Public DGS Corpus\citep{hankeExtendingPublicDGS2020} &DGS&\cmark &\xmark&\xmark & \xmark   &50  &-  & -         &TV    \\  
  PHOENIX-News (ours)                &DGS&\cmark &\cmark&\cmark &\cmark &486  &190k & 11        &TV   \\ 
  \hline
  \end{tabular}
  }
\end{table*}

In this work, we are inspired by recent advances from learning the discrete representation for generation \citep{huangAccurateImageCoding2023, zhangT2MGPTGeneratingHuman2023,williams2020hierarchical,vandenoordNeuralDiscreteRepresentation2017,ao2022rhythmic}.  Specifically, we investigate a two-stage framework based on Dynamic Vector Quantized Variational Autoencoders (DVQ-VAE) and Generative Pre-trained Transformer (GPT) \citep{radford2018improving} for text-to-sign language production.
In the first stage, as shown in \autoref{fig:introduction}, DVQ-VAE will learn the weights of each frame and the boundaries of the basic semantic units.
Then, the weighted latent vectors are mapped to discrete code indices.
Further quantitative analysis of the uneven information density in sign language is provided in \autoref{sec:Analyzing}.
To encourage models to perform variable-length encoding and compress sequence lengths, we propose a novel budget loss.
Additionally, to preserve the semantic information of the reconstructed sign language sequences, we also introduce a translation auxiliary loss.
In the second stage, a GPT-like model is learned to to generate code index sequences from spoken language text.
Furthermore, since the duration of quantized code in a sequence can also vary dynamically, we further propose a duration transformer to predict the duration of the next code based on the previous code's duration and the current code.

The experimental results on the widely used SLP dataset PHOENIX14T \citep{camgozNeuralSignLanguage2018} demonstrate that our proposed method achieves superior back translation performance compared to previous approaches.
Furthermore, throughout the entire development process of image generation and text generation, the scale of the dataset has played a crucial role.
A large amount of high-quality corpus is also very important for SLP tasks.
In this paper, we present the largest known German Sign Language dataset, PHOENIX-News, which consists of 486 hours of sign language videos, audio, and transcription texts.
The native expression, clear hand details, and extensive coverage of our large-scale dataset make it suitable for a variety of sign language research tasks, such as sign language translation and sign language production.
Based on this dataset, we further explore the impact of training data size on SLP tasks.
Empirical analysis shows that the performance of our model can be further improved by increasing the size of the training data.

Our main contributions are summarized as follows:

\begin{itemize}
  \item We analyse the uneven information density in sign language. Additionally,we propose for the first time an information density based variable length coding method suitable for sign language.
  \item We propose a two-stage SLP framework consisting of two components: 1) DVQ-VAE to dynamically assign variable-length codes to sequences based on their different information densities through a novel \emph{adaptive downsampling module} and \emph{budget loss}. 2) A novel \emph{T2M-GPT model} to predict variable-length codes and their corresponding durations.
  \item Extensive experiments on the challenging PHOENIX14T dataset show the effectiveness of our proposed method.
  \item We propose the largest known German sign language dataset, PHOENIX-News, which can be used for a variety of sign language research tasks.

\end{itemize}

\section{Related Work}
\subsection{Sign Language Production}
Sign language production (SLP) has been an active area of research for nearly two decades\citep{cox2002tessa,mcdonald2016automated}. 
Early approaches focused on mapping text to glosses using neural models. \citep{stollText2SignSignLanguage2020a} proposed a seq2seq architecture for SLP, which mapped text input to glosses. To generate 2D joint locations, they utilized an empirical lookup table paradigm.
Then, \citep{saundersProgressiveTransformersEndtoend2020} proposed a progressive transformer to directly learn the mapping between annotations and skeleton pose sequences.
\citep{saundersAdversarialTrainingMultichannel2020} proposed to improve the quality of skeleton pose generation through adversarial training.
In addition, several approaches have been proposed to enhance the generation quality through the utilization of mixture density networks\citep{saundersContinuous3DMultichannel2021}, Mixture-of-Experts\citep{saundersMixedSIGNalsSign2021}, dictionary representations\citep{saundersSigningScaleLearning2022}, and diffusion models\citep{baltatzis2023neural}.
Several studies have proposed the use of non-autoregressive models to generate sign language, thereby improving generation speed \citep{huangFastHighQualitySign2021, hwangNonautoregressiveSignLanguage2021, hwangNonAutoregressiveSignLanguage2022}.
Additionally, researchers have explored the generation of photo-realistic sign language videos using Generative Adversarial Networks (GANs) \citep{saundersSigningScaleLearning2022} or diffusion models \citep{fangSignDiffLearningDiffusion2023, xieSignLanguageProduction2024}.
Recent studies have shown that using 3D human models, such as SPML-x\citep{pavlakosExpressiveBodyCapture2019}, is a better choice for sign language understanding \citep{leeHumanPartwise3D2023} and production tasks \citep{stollThereBackAgain2022}.
\citep{inan2022modeling} found that representing the intensification level of glosses connected with the duration of a sign.

\subsection{Vector Quantization for SLP}
Vector Quantized Variational Autoencoders (VQ-VAE) proposed by \citep{vandenoordNeuralDiscreteRepresentation2017} is an autoencoder structure that aims to learn a discrete representation of data.
Recently, VQ-VAE has been used for the SLP task, such as \citep{saundersContinuous3DMultichannel2021} using a modified VQ-GAN for isolated word sign language video generation.
Recently, VQ-VAE has been applied to the SLP task. For instance, \citep{xieSignLanguageProduction2024} utilized a modified VQ-GAN \citep{esserTamingTransformersHighResolution2021} to generate isolated sign language videos.
\citep{xieG2PDDMGeneratingSign2023} employed VQ-VAE to generate sign pose sequences from gloss sequences.
However, existing methods rely on fixed-length encodings and overlook the unequal distribution of information in sign language. To address this issue, we propose a pioneering approach: a variable-length dynamic vector quantization method specifically designed for sign language.

\subsection{Sign Language Dataset}
High-quality sign language datasets are crucial for the SLP task.
\autoref{tab:dataset} summarizes the publicly available datasets used for sign language research.
The PHOENIX14T \citep{camgozNeuralSignLanguage2018} dataset is the most commonly used dataset for SLP tasks, but it has limited data.
As an important supplement, we propose PHOENIX-News, which contains 486 hours of sign language data. To the best of our knowledge, this is the largest German sign language dataset to date.

\begin{figure}[htbp]
  \subcaptionbox{Length distribution of different glosses. \label{fig:duration_hist}}{\includegraphics[width=0.49\linewidth]{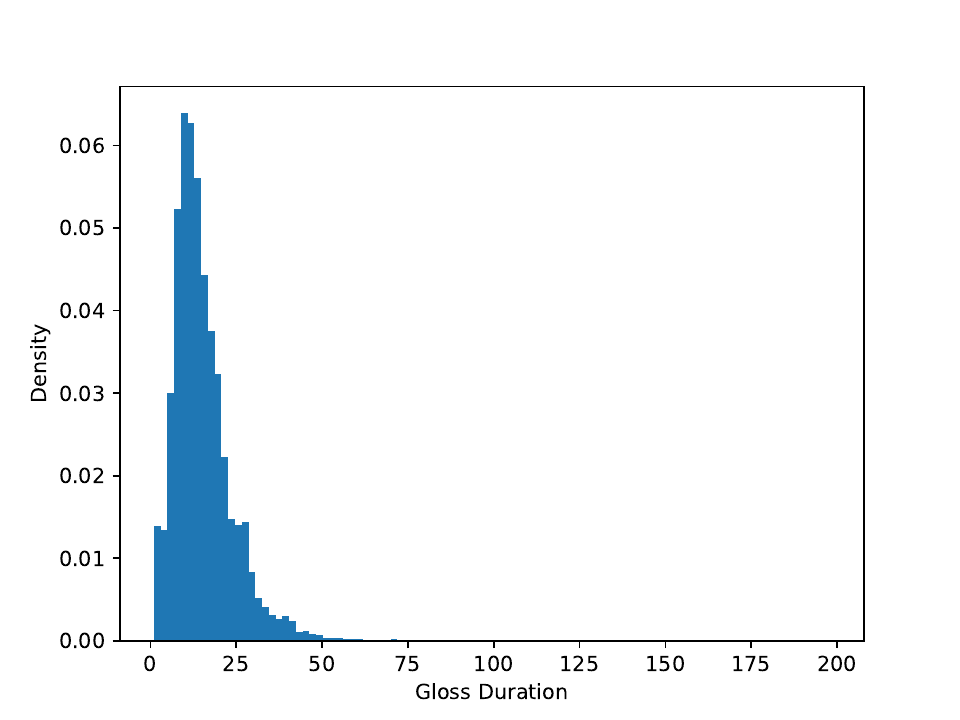}}
  \subcaptionbox{Length distribution of the same gloss. \label{fig:duration_hist_most_common}}{\includegraphics[width=0.49\linewidth]{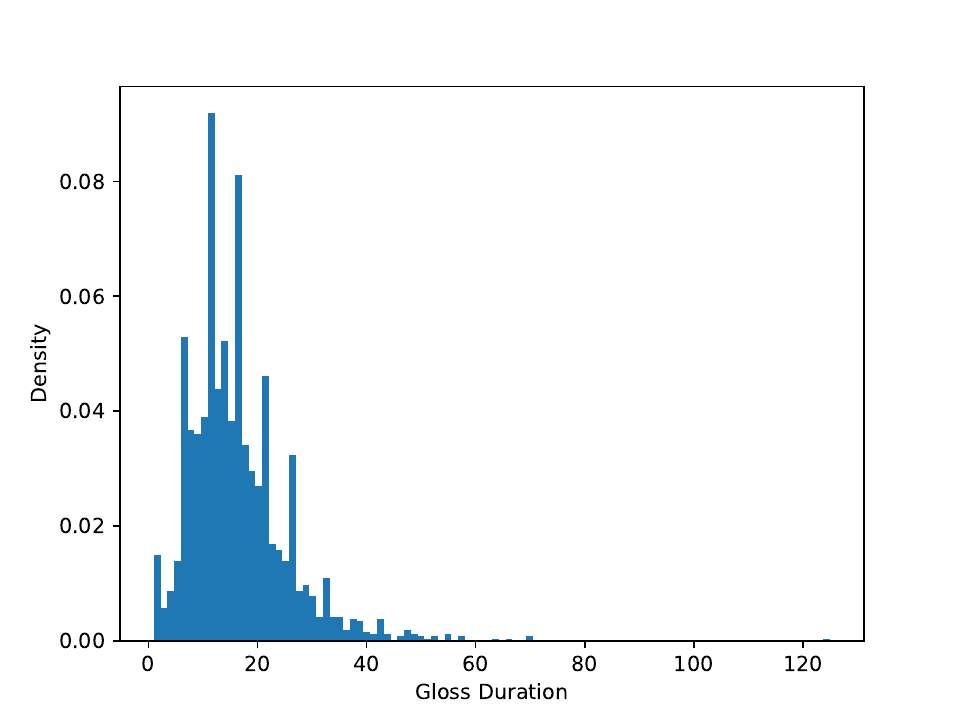}}
  \centering
\end{figure}
\section{Analyzing Information Density in Sign Language}
The most commonly used discrete representation for sign language is glosses, which are the basic semantic units in sign language and are annotated by sign language experts.
We first counted the length distribution of glosses in the PHOENIX14T dataset, as shown in \autoref{fig:duration_hist}.
It can be seen that the length distribution of glosses is uneven, with most glosses having a length between 0 and 50, but some glosses have a length of more than 50.
This indicates that uneven information density does exist in sign language.
We then counted the length distribution of the most frequently occurring glosses (REGEN) in different contexts, as shown in \autoref{fig:duration_hist_most_common}.
It can be seen that even the same gloss has different lengths in different contexts.
These analysis results inspire us to design a dynamic vector quantization method as described in \autoref{sec:dvq-vae} and a duration transformer to predict duration based on context as described in \autoref{sec:t2s-gpt}.
\label{sec:Analyzing}

\begin{figure*}[htp]
  \centering
  \includegraphics[width=0.9\linewidth]{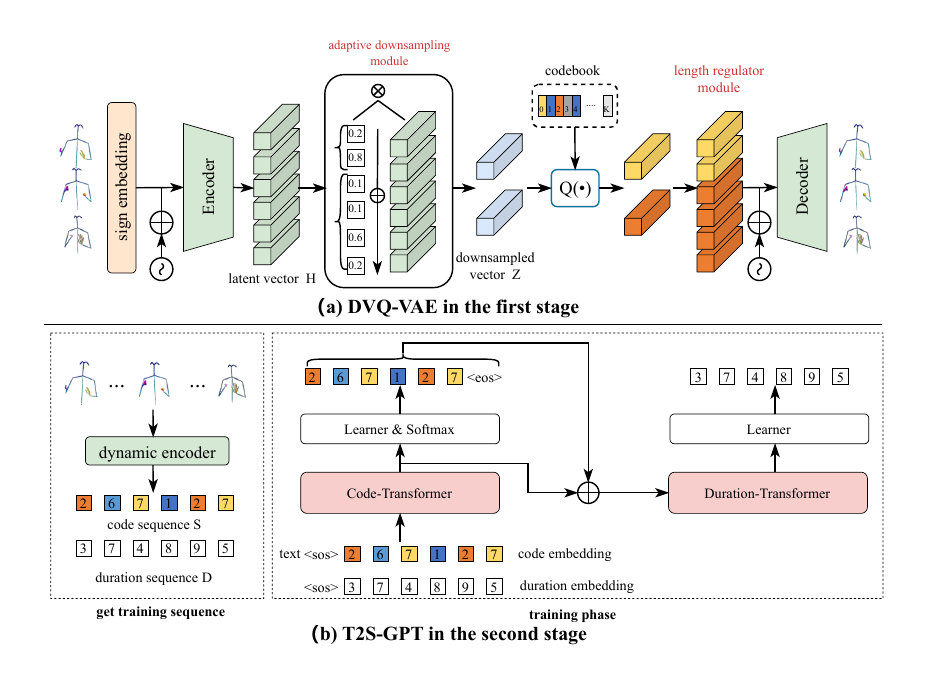}
  \caption{The overview of our proposed two-stage framework.}
  \label{fig:method}
\end{figure*}

\section{Method}
\label{sec:method}
Our overall two-stage framework is depicted in \autoref{fig:method}, which consists of two stages: DVQ-VAE and T2M-GPT.
In the following, we will first briefly revisit the formulation of VQ and then describe our proposed method in detail.

\subsection{Preliminary}
Vector quantization (VQ) \citep{vandenoordNeuralDiscreteRepresentation2017} represents a technique for learning a codebook to encode sign language sequences into discrete code representations.
Given a sign language sequence $X=\left[x_1, x_2, \ldots, x_T\right]$ with $x_t \in \mathbb{R}^d$, where $T$ is the number of frames and $d$ is the dimension of the sign language, we aim to recover the sign language sequence through an autoencoder and a learnable codebook containing $K$ codes $C=\left\{c_k\right\}_{k=1}^K$ with $c_k \in \mathbb{R}^{d_c}$, where $d_c$ is the dimension of codes.
The sign language sequence $X$ is first encoded by the encoder $E$ into a sequence of latent vectors $Z=\left[z_1, z_2, \ldots, z_{T/l}\right]$, and $z_t \in \mathbb{R}^{d_c}$, where $l$ represents the temporal downsampling rate of the encoder $E$.
For fixed-length encoding, $l$ is fixed, while for variable-length encoding, $l$ is dynamically changing.
For $i$-th latent feature $z_i$, the quantization through $C$ is to find the most similar element in $C$, which can be properly written as:

\begin{equation}
  \hat{z_i} =  \underset{c_k\in C}\arg\min\|z_i - c_k\|_2
\end{equation}

\subsection{Stage 1: Dynamic Vector Quantization VAE (DVQ-VAE)}
\label{sec:dvq-vae}
Existing methods employ a fixed downsampling rate $l$ for fixed-length encoding, neglecting the uneven information density in sign language. This oversight introduces redundancy in the learned codebook, leading to a decrease in both generation quality and speed.
To address this issue, we propose DVQ-VAE, which consists of a dynamic encoder and a dynamic decoder.

\paragraph{Dynamic Encoder.}
As shown in \autoref{fig:method}, the sign language sequence $X$ first passes through a sign language embedding layer. Then, after adding positional encoding information, it is input into a Transformer Encoder to obtain a sequence of latent vectors $H=\left[h_1, h_2, \ldots, h_T\right]$, with $h_t \in \mathbb{R}^{d_h}$, where $d_h$ is the dimension of the latent vectors.
We formulate these operations as:
\begin{equation}
  \begin{split}
    X'_t =& relu(LN(W_1X_t + B_1)) + f_{pos}(t) \\
    H =& \text{TransformerEncoder}(X')
  \end{split}
\end{equation}
where LN denotes Layer Normalization \citep{baLayerNormalization2016}, $f_{pos}$ denotes the positional encoding function, and $W_1 \in \mathbb{R}^{d_h \times d}$ and $B_1 \in \mathbb{R}^{d_h}$ are learnable parameters.

The dynamic encoder then contains an \textbf{information-based adaptive downsampling module}, which adaptively adjusts the downsampling rate by considering the information weight of each frame.
Specifically, we input the latent vector sequence $H$ into a multi-layer perceptron (MLP) to obtain the information weight of each frame $I=\left[i_1, i_2, \ldots, i_T\right]$, where $i_t \in [0, 1]$.
We then segment the latent vector sequence $H$ according to the information weight threshold $O$ (we set it to 1.0) for semantic unit, and then perform weighted averaging within the segment to obtain the downsampled latent vector sequence $Z=\left[z_1, z_2, \ldots, z_{T/l}\right]$.
The downsampling process of the entire module can be formulated as:
\begin{equation}
  I = \sigma (W_3(relu(W_2H+B_2)+H)+B_3)
  \label{eq:mlp}
\end{equation}
\begin{equation}
  S = cumsum(I) // O
  \label{eq:segment}
\end{equation}
\begin{equation}
  \begin{split}
    &Z_t = \sum_{j=1}^T H_j \cdot I_j \cdot F_j ,  \quad\text{D}_t = \sum_{j=1}^T \cdot F_j \\
    &\text{where } F_j= \begin{cases}
      1, & \text{if } S_j = t-1 \\
      0, & \text{otherwise}
    \end{cases}
  \end{split}
  \label{eq:downsample}
\end{equation}
\autoref{eq:mlp} represents the operation of the MLP, where $\sigma$ denotes the sigmoid activation function.
\autoref{eq:segment} represents the process of segmenting the latent vector sequence $H$ according to the information weight threshold $O$, where $S = \left[s_1, s_2, \ldots, s_{T}\right]$ and $s_t \in [0, sum(I)//O]$, representing the position markers of the segments.
$cumsum$ denotes the cumulative sum function, and // denotes the integer division.
\autoref{eq:downsample} represents the process of weighted downsampling, where $Z_t$ denotes the downsampled latent vector and $D_t$ denotes the duration of the current latent vector.
\paragraph{Dynamic Decoder.}
The goal of the dynamic decoder is to reconstruct the original sign language sequence $X$ based on the quantized latent vector sequence $\hat{Z}$ and the duration information $D=\left[d_1, d_2, \ldots, d_{T/l}\right]$.
We use a \textbf{length regulator module} to address the issue of mismatched lengths between the vector sequence $\hat{Z}$ and the original sign language sequence $X$ during dynamic decoding.
\begin{equation}
  \hat{X} = \text{LR}(\hat{Z}, D)
  \label{eq:lr}
\end{equation}
where $\hat{X}$ denotes the extended sequence, and LR denotes the length regulator module.
For example, if $\hat{Z}=[\hat{z}_1, \hat{z}_2, \hat{z}_3]$ and $D=[1, 2, 3]$, then $\hat{X}=[\hat{z}_1, \hat{z}_2, \hat{z}_2, \hat{z}_3, \hat{z}_3, \hat{z}_3]$.
We then input the extended sequence $\hat{X}$ into a Transformer-based decoder to obtain the reconstructed sign language sequence $X_{re}$.
\paragraph{Training of DVQ-VAE.}
The optimization goal of the original VQ-VAE \citep{vandenoordNeuralDiscreteRepresentation2017} $\mathcal{L}_{\mathrm{vq}}$ contains three components: a reconstruction loss $\mathcal{L}_{\text {re }}$, an embedding loss $\mathcal{L}_{\text {embed }}$, and a commitment loss $\mathcal{L}_{\text {commit }}$.
\begin{equation}
  \mathcal{L}_{v q}=\mathcal{L}_{r e}+\underbrace{\|Z-s g[\hat{Z}]\|_2}_{\mathcal{L}_{\text {embed }}}+\lambda_1 \underbrace{\|s g[Z]-\hat{Z}\|_2}_{\mathcal{L}_{\text {commit }}}
\end{equation}
where $\lambda_1$ is a hyper-parameter for the commitment loss and $s g$ is the stop-gradient operator.
\begin{figure*}[htbp]
  \centering
  \subcaptionbox{Video duration (seconds) \label{fig:video_duration} }{\includegraphics[width = 0.49\linewidth]{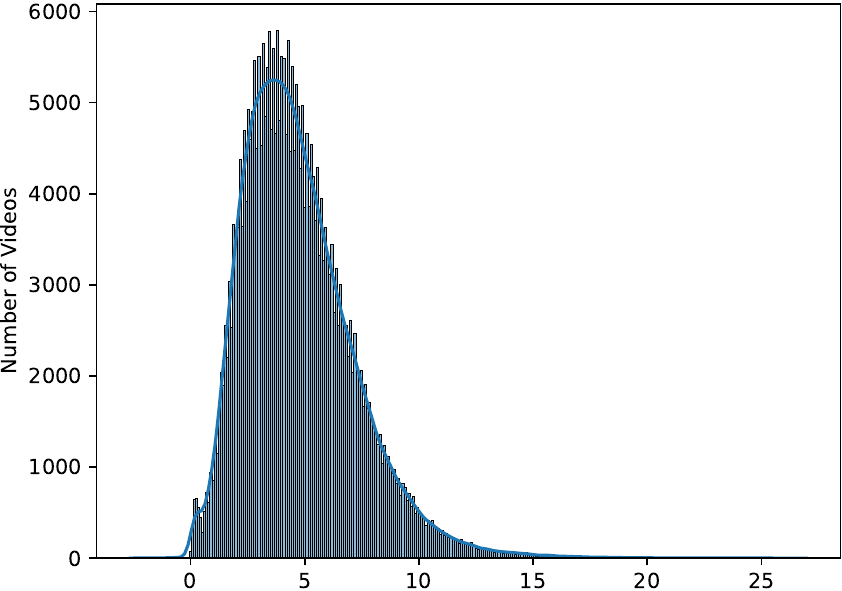}}
  \subcaptionbox{Text length \label{fig:text_len}}{\includegraphics[width = 0.49\linewidth]{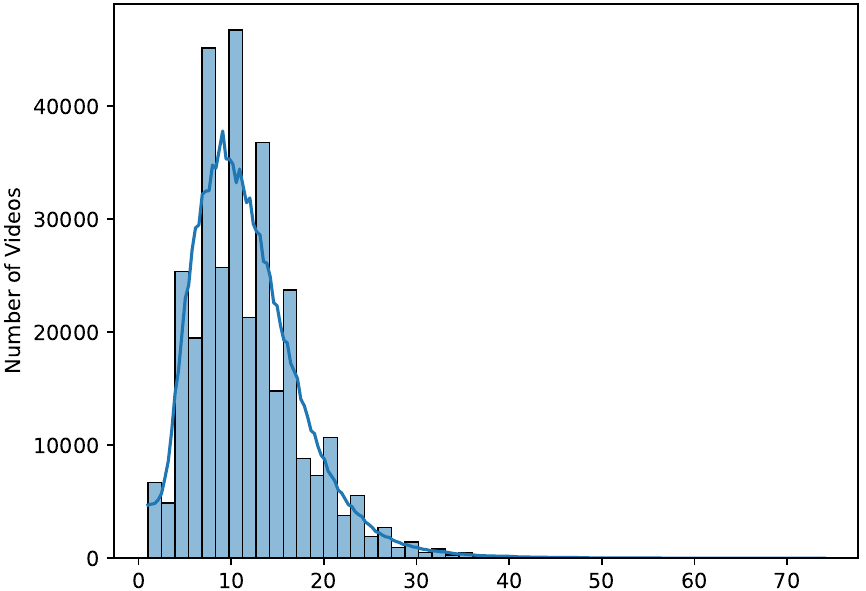}} 
  \caption{Distribution of text length and video duration in the PHOENIX-News dataset.}
\end{figure*}
In our work, the calculation formula for the reconstruction loss is as follows:
\begin{equation}
  \mathcal{L}_{r e}=\mathcal{L}_1^{\text {smooth }}\left(X, X_{\text {re }}\right)+ \mathcal{L}_1^{\text {smooth }}\left(V(X), V\left(X_{\text {re }}\right)\right)
\end{equation}
where $V(\cdot)$ denotes the calculation of velocity, for example, $V(X) = [v_1, v_2, \ldots, v_{T-1}]$, where $v_i=x_{i+1}-x_i$.
In addition to the original optimization goal, we introduce two new loss functions: \textbf{budget loss} $\mathcal{L}_{\text {budget}}$ and \textbf{sign language translation auxiliary loss} $\mathcal{L}_{\text {slt}}$. 
Without using the budget loss, the model tends to use more codes to represent the sign language sequence, resulting in longer sequence lengths.
To encourage the model to use a higher downsampling rate $l$, we define the budget loss as:
\begin{equation}
  \mathcal{L}_{\text {budget}} = \mathbb{E} [max(0,(sum(I)-T/R))]
\end{equation}
Since the length of the downsampled sequence is $sum(I)//O$, the budget loss can be interpreted as the expectation of the length of the downsampled sequence.
Where $T$ denotes the length of the original sign language sequence, and $R$ denotes the expected downsampling rate.
The goal of the sign language translation auxiliary loss is to preserve the semantic information of the reconstructed sign language sequence, and its calculation formula is as follows:
\begin{equation}
  \mathcal{L}_{\text {slt}} = \mathbb{E} [-log P(Y|X_{re})]
\end{equation}
where $Y$ denotes the spoken language text corresponding to the sign language sequence.
The final loss for DVQ-VAE is defined as:
\begin{equation}
  \mathcal{L} = \mathcal{L}_{v q} + \lambda_2 \mathcal{L}_{\text {budget}} + \lambda_3 \mathcal{L}_{\text {slt}}
\end{equation}
We also use two common training recipes \citep{razavi2019generating}, exponential moving average (EMA) and code book restart, to improve the utilization of the codebook.

\begin{table*}[htp]
  \caption{Quantitative results for text to sign language task on PHOENIX14T test set. }
  \label{tab:main_results}
  \small
  \centering
  \begin{tabular}{lccccc}
    \toprule
    \textbf{Methods}  & \textbf{ROUGE-L} &\textbf{ BLEU-1} & \textbf{BLEU-2} & \textbf{BLEU-3} & \textbf{BLEU-4} \\
    \midrule
    GT &39.17 & 37.75 & 24.92 & 18.25 & 14.34 \\
    \midrule
    PT\citep{saundersProgressiveTransformersEndtoend2020} &20.58 & 17.47 & 7.76 & 5.50 & 4.38 \\
    NAT-EA\citep{huangFastHighQualitySign2021} &26.81 & 27.00 & 14.12 & 9.20 & 6.67 \\
    \midrule
    T2M-GPT~\citep{zhangT2MGPTGeneratingHuman2023}  &29.19 & 28.32 & 16.05 & 10.77 & 8.01 \\
    MDM~\citep{tevetHumanMotionDiffusion2022}  &30.37 &27.59 & 15.83 & 10.29 & 7.55 \\
    \midrule
    \textbf{T2S-GPT (ours)}  &\textbf{34.65} & \textbf{33.16} & \textbf{21.09} & \textbf{15.26} & \textbf{11.87} \\
    \bottomrule
  \end{tabular}
\end{table*}

\subsection{Stage 2: Text-to-Sign GPT (T2S-GPT)}
\label{sec:t2s-gpt}
\paragraph{Code-Transformer.}
With a learned DVQ-VAE, a sign language sequence $X$ can be mapped to a sequence of indices $S=$ $\left[s_1, s_2, \ldots, s_{T / l}, E n d\right]$, which are indices from the learned codebook.
Note that a special $End$ token is added to indicate the stop of the sign language code sequence.
By projecting $S$ back to their corresponding codebook entries, we obtain $\hat{Z}=\left[\hat{z}_1, \hat{z}_2, \ldots, \hat{z}_{T / l}\right]$, where $\hat{z}_i=c_{s_i}$.
The generation of the sign language code sequence $S$ can be formalized as an autoregressive next index prediction problem: given the previous $i-1$ indices, i.e., $S_{<i}$, and the text condition $Y$, our goal is to predict the distribution of the possible next index $p\left(S_i \mid Y, S_{<i}\right)$, which can be solved by a transformer, as shown in \autoref{fig:method}.
The negative log-likelihood (NLL) loss for code autoregressive training is:
\begin{equation}
  \mathcal{L}_{\text {code}} = \mathbb{E} [ -\log p\left(S_i \mid Y, S_{<i},D_{<i}\right)]
\end{equation}
We introduce a duration embedding layer to embed the duration information $D$ into the transformer.
\paragraph{Duration-Transformer.}
As mentioned in \autoref{sec:dvq-vae} and \autoref{eq:lr}, to decode $X_{re}$, we need not only $\hat{Z}$, but also the duration information $D$.
Therefore, we design a duration-transformer to predict the duration of the next code based on the previous code's duration and the current codes.
As shown in \autoref{fig:method}, the duration-transformer takes the sum of Code-Transformer's output hidden vector $H_{code}$ and an extra code embedding as input:
\begin{equation}
  \begin{split}
  H_{dur} = &H_{code}[N_{y}:N_y+l-1] \\
   &+ f_{code}(S[\leq l])
  \end{split}
\end{equation}
where $H_{dur}$ denotes the input of the duration-transformer, and $N_y$ denotes the length of the condition text.
The design idea behind this is that when predicting the next code's duration, the model should not only be aware of previous steps'codes and their duration information but also should be aware of current code information.
The optimization goal of the duration-transformer is to minimize the difference between the predicted duration and the real duration.
The calculation formula is as follows:
\begin{equation}
  \mathcal{L}_{\text {dur}} = \mathbb{E} [ \|D_i - \hat{D}_i\|_2]
\end{equation}
In inference, we round the output of the duration-transformer to obtain the duration.
The final optimization goal is:
\begin{equation}
  \mathcal{L} = \mathcal{L}_{\text {code}} +  \mathcal{L}_{\text {dur}}
\end{equation}

\section{The Proposed PHOENIX-News Dataset}
As shown in \autoref{tab:dataset}, PHOENIX-News aims to provide the community with a new large-scale document-level sign language dataset, which contains 486 hours of sign language videos, audio, and transcription texts.
We collected daily news programs in German Sign Language from the German public television station PHOENIX from 2013 to 2023.
We then used whisper \citep{radford2023robust} to transcribe the program's speech into text.
Finally, we performed preprocessing steps such as domain cropping, sign language pose estimation, and sign language text alignment to obtain the final dataset.
Since there are new sign language news programs every day, PHOENIX-News will be a continuously updated project. 
Therefore, we did not divide the dataset into training and test sets, but used the existing dataset for testing.
Each video in the dataset has an average duration of 4.7 seconds, and the average text length is 11 words.
We show the distribution of video duration and text length in \autoref{fig:video_duration} and \autoref{fig:text_len}.

\begin{figure*}[htbp]
  \centering
  \includegraphics[width=0.7\linewidth]{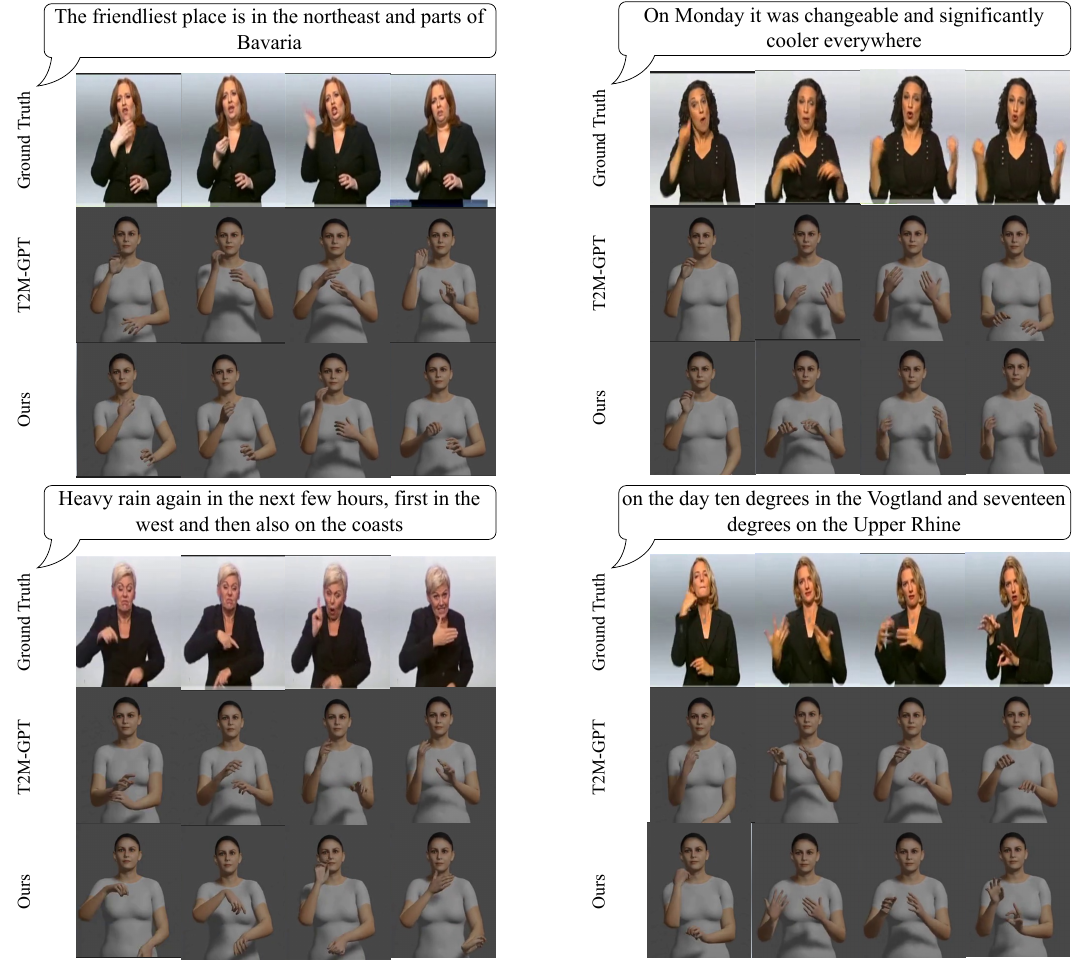}
  \caption{Qualitative results of our T2S-GPT model.}
  \label{fig:qualitative_results}
\end{figure*}

\section{Experiments}
\subsection{Experimental Setup}
We will introduce our experimental setup in this section, including the dataset and evaluation metrics.
We will provide the implementation details of the model in the \autoref{appe:implementation}.
\paragraph{Dataset and Evaluation Metrics.}
We evaluate our proposed T2S-GPT model on the PHOENIX14T \citep{camgozNeuralSignLanguage2018} dataset, which is the most commonly used dataset for SLP tasks and has been used as a benchmark in many previous SLP works \citep{huangFastHighQualitySign2021, saundersContinuous3DMultichannel2021, saundersProgressiveTransformersEndtoend2020, xieG2PDDMGeneratingSign2023}.
The PHOENIX14T dataset contains 7,096 training samples (with 2,887 words in German spoken language translations), 519 validation samples, and 642 test samples.
We use the pose parameter $\theta$ in the SMPL-X model to represent the sign language pose, and the rotation 6D representation \citep{zhouContinuityRotationRepresentations2019a} is used to represent the rotation in the pose.
Following the most widely used setting in SLP \citep{saundersProgressiveTransformersEndtoend2020}, we use the back translation metric to evaluate the generation quality.
Since previous works did not publicly release the weights of their SLT models used to calculate the back translation metric, following the previous setting, we train an SLT model using the code from \citep{camgozSignLanguageTransformers2020}.
We provide the details of the sign language representation in the \autoref{appe:sign-representation}.
We provide the training details of the SLT model in the \autoref{appe:slt}.

\subsection{Comparisons with State-of-the-Art Methods}
We compare our T2S-GPT model with several other models, including the state-of-the-art text-to-sign model and the state-of-the-art text-to-motion model.
\paragraph{Comparison methods.}
1) Progression Transformer (PT) \citep{saundersProgressiveTransformersEndtoend2020} directly predicts the sign language pose sequence in an autoregressive manner.
2) NAT-EA \citep{huangFastHighQualitySign2021} generates the sign language pose sequence in a non-autoregressive manner.
3) T2M-GPT \citep{zhangT2MGPTGeneratingHuman2023} is a state-of-the-art autoregressive text-to-motion model, and its prediction target is the discrete representation of sign language processed by VQ-VAE.
4) MDM \citep{tevetHumanMotionDiffusion2022} uses a diffusion model to generate motion sequences based on text in a non-autoregressive manner.
Both T2M-GPT and MDM use CLIP \citep{radford2021learning} to extract text features as condition signals, but the original CLIP does not support German well. To make a fair comparison, we use a multilingual CLIP model\citep{reimers-2019-sentence-bert}\footnote{\url{https://huggingface.co/sentence-transformers/clip-ViT-B-32-multilingual-v1}}.

\paragraph{Quantitative Comparison.}
We report the back translation metrics (including BLEU \citep{papineni2002bleu} scores and ROUGE-L \citep{lin2004automatic} scores) obtained by all models on the PHOENIX14T dataset in \autoref{tab:main_results}.
We only use the PHOENIX14T dataset to train all models, and the training settings are consistent with those in the original papers.
As shown in \autoref{tab:main_results}, our T2S-GPT model achieves the best results on all metrics.
Specifically, our T2S-GPT model achieves a score of 11.87 on BLEU-4, which is 3.86 points higher than the state-of-the-art T2M-GPT model.
On ROUGE-L, our T2S-GPT model achieves a score of 34.65, which is 4.28 points higher than the state-of-the-art MDM model.
These results indicate that our T2S-GPT model can generate higher-quality sign language.

\paragraph{Qualitative Results.}
We qualitatively compare our T2S-GPT method with other methods and the ground truth sign language pose sequence on the PHOENIX14T test set, as shown in \autoref{fig:qualitative_results}.
As shown in \autoref{fig:qualitative_results}, compared with other methods, the sign language generated by our T2S-GPT method is closer to the ground truth.
Note that we provide a video demonstration of our method on the anonymous project homepage \url{https://t2sgpt-demo.yinaoxiong.cn/}, which can better convey the temporal information.
\begin{table}[htb]
  \centering
  \small
  \caption{Results of ablation experiments on the PHOENIX14T dataset.}
  \label{tab:ab}
  \resizebox{\linewidth}{!}{
  \begin{tabular}{lccccc}
    \toprule  %
    \textbf{Model}&\textbf{R}&\textbf{B1}&\textbf{B2}&\textbf{B3}&\textbf{B4}\\
    \midrule  %
    T2S-GPT              &\textbf{34.65} & \textbf{33.16} & \textbf{21.09} & \textbf{15.26} & \textbf{11.87} \\
    w/o DVQ-VAE         &30.80 & 27.77 & 16.01 &10.96 & 8.39 \\
    w/o Duration-Transformer         &31.99 & 30.05 & 18.25 &12.43 & 9.39 \\
    \bottomrule %
    \end{tabular}
  }
    \label{table:ab}
\end{table}

\subsection{Ablation and Analysis}

\textbf{Analysis on DVQ-VAE.}
As shown in the second row of \autoref{tab:ab}, when we replace DVQ-VAE with the VQ-VAE proposed by \citep{zhangT2MGPTGeneratingHuman2023} with a downsampling rate of 4, we find that the back translation metrics of the SLP model have decreased significantly.
This indicates that our DVQ-VAE model can obtain more compact and higher-quality discrete representations of sign language.

\textbf{Analysis on Duration-Transformer.}
As shown in the third row of \autoref{tab:ab}, when we replace the duration-transformer with a simple fully connected layer, we find that the back translation metrics of the SLP model have decreased significantly.

\begin{figure}[htbp]
  \centering
  \includegraphics[width=0.8\linewidth]{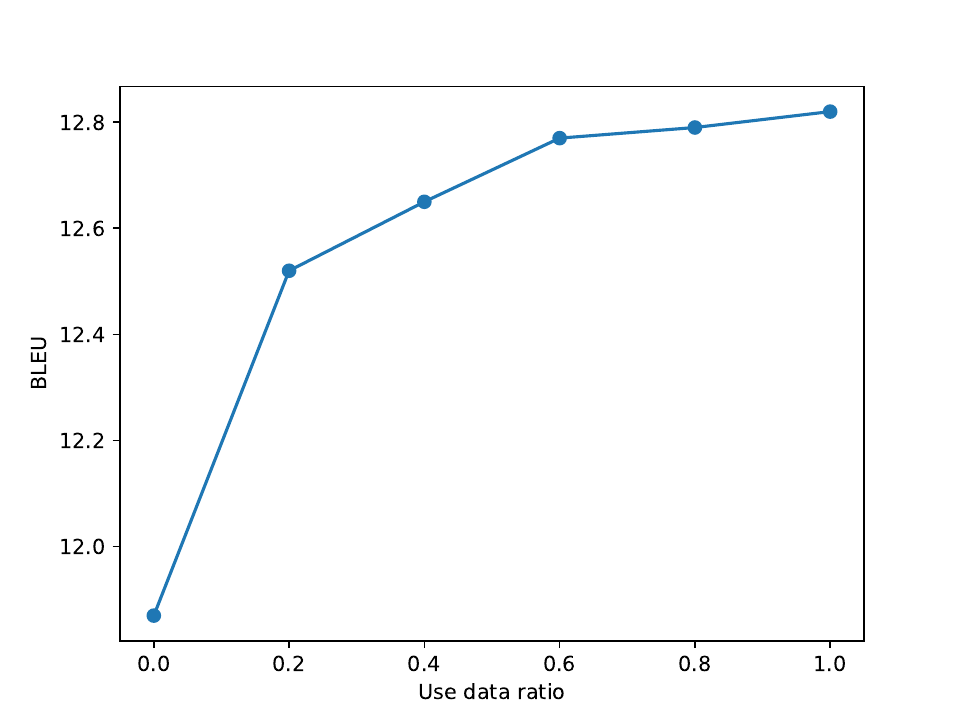}
  \caption{Impact of dataset size on the performance of T2S-GPT.}
  \label{fig:scale}
\end{figure}

\textbf{Impact of dataset size.}
To study whether our proposed T2S-GPT model is scalable, we train the T2S-GPT model by adding different proportions of the PHOENIX-News dataset.
The experimental results are shown in , where we find that as the dataset size increases, the performance of the T2S-GPT model also continues to improve.
This indicates that our T2S-GPT model is scalable.

\section{Conclusion}
In this work, we propose a two-stage text-to-sign model T2S-GPT, which consists of a dynamic vector quantization VAE (DVQ-VAE) and a GPT-like autoregressive generation model.
Our method achieves better performance than the previous state-of-the-art text-to-sign model.
In addition, we have collected a new large-scale document-level sign language dataset PHOENIX-News, and the experimental results show that a larger dataset can still bring additional improvements to our method.
\section{Limitations and Potential Risks}
Although using a 3D human body model as the sign language representation introduces prior information about human body shape, it does not constrain the rotation motion of the joints themselves.
The model's predictions occasionally produce some abnormal cases that do not conform to the human joint structure, which may make users feel uncomfortable.
At the same time, this is also a manifestation of the model's generation errors.
To address this issue, we plan to introduce more prior information in future work, such as human motion priors and physical constraints on human joint rotation angles.
SLP technology itself does not have any obvious potential risks, but since the current SLP technology is still in a relatively early stage, if it is directly applied to practical scenarios, it may mislead users.
For example, in weather forecasts, if the model generates sign language with incorrect place names, it may mislead users.

\section*{Acknowledgements}

This work was supported by the Key Research and Development Projects in Zhejiang Province (No. 2024C01106), the NSFC (No. 62272411), the National Key Research and Development Project of China (2018AAA0101900),  and Research funding from FinVolution Group.

\bibliography{custom,sign-codec}

\appendix

\section{Example Appendix}
\label{sec:appendix}

\section{Sign Language Representation}
\label{appe:sign-representation}
Inspired by the latest advances in sign language processing \citep{leeHumanPartwise3D2023, stollText2SignSignLanguage2020a}, to better represent the complex body movements in sign language, we propose to use the pose parameter $\vec{\theta}=\left[\vec{\omega}_0^T, \ldots, \vec{\omega}_K^T\right]^T$ of the SMPL-X human body model \citep{pavlakosExpressiveBodyCapture2019} as the sign language representation, instead of the 3D joint coordinates in Euclidean space used in previous works.
Where $\vec{\omega}_k \in \mathbb{R}^3$ denotes the axis-angle representation of the relative rotation of part $k$ with respect to its parent in the kinematic tree.
However, since the axis-angle form is not a continuous rotation representation, which is not conducive to network learning, we further convert it to the rotation 6D representation \citep{zhouContinuityRotationRepresentations2019a} $\vec{o  }=\left[\vec{r}_0^T, \ldots, \vec{r}_K^T\right]^T$.
We ignore the lower body joints outside the visible range.
There are three advantages of using this representation: 1) it has rotation and translation invariance; 2) it separates the modeling of human body shape and pose, and the semantics of sign language should only be related to the pose and independent of the shape; 3) the introduction of human body prior avoids generating abnormal results, such as fingers longer than arms.

\section{Implementation Details}
\label{appe:implementation}
For DVQ-VAE, we set the dimension of the latent vectors $d_h$ to 512, the dimension of the codebook $d_c$ to 512, and the number of codes $K$ to 1024.
The number of transformer layers in the encoder and decoder is set to 6, and the hidden size, number of heads, and feed-forward dimension for each layer are set to 512, 8, and 2048, respectively. The dropout rate is set to 0.1.
We use AdamW \citep{loshchilov2018decoupled} optimizer with $\left[\beta_1, \beta_2\right]=[0.9,0.99]$, batch size of 256 , and exponential moving constant $\lambda=0.99$.
We train for a total of 100K iterations, with an initial learning rate of 2e-4, and then use the cosine learning rate decay strategy during training.
$\lambda_1$, $\lambda_2$, and $\lambda_3$ in the final loss are set to 1, 0.5, and 1.0, respectively.
The $R$ in $\mathcal{L}_{budget}$ is set to 12.

For T2S-GPT, the hidden size, number of heads, and feed-forward dimension for each transformer layer are set to 1024, 16, and 4096, respectively. The dropout rate is set to 0.1.
The number of transformer layers in the code-Transformer and duration-Transformer is set to 18 and 6, respectively.
We use a batch size of 256 and train for 300K iterations.
We optimize the models with the AdamW optimizer, warm up the learning rate for the first 4k updates to a peak of 1e-4, and then linearly decay it to 0.
We use a 32GB NVIDIA V100 GPU to train our model.

\section{Back Translation Model}
\label{appe:slt}
To calculate the back translation metric, we train a sign language translation (SLT) model that takes sign language pose sequences as input and outputs the corresponding spoken language text.
The SLT model adopts the architecture introduced by \citep{camgozSignLanguageTransformers2020}. Both the encoder and decoder components of the model are built using transformers. In particular, the hidden size, number of heads, and feed-forward dimension for each layer are configured as 512, 8, and 2048, respectively. Additionally, a dropout rate of 0.4 is applied within the model.
The number of transformer layers in the encoder and decoder is set to 3.
The training settings are consistent with those in the original paper.

\end{document}